\documentclass[runningheads,a4paper]{llncs}

\usepackage{amssymb}
\usepackage{graphicx}

\usepackage{amsmath}
\usepackage{bbm}
\usepackage{algorithm}
\usepackage{algpseudocode}

\usepackage{url}
\urldef{\mailsc}\path|{francis.maes, L.Wehenkel, dernst}@ulg.ac.be|    
\newcommand{\keywords}[1]{\par\addvspace\baselineskip
\noindent\keywordname\enspace\ignorespaces#1}

\newcommand{\smallparagraph}[1]{\textit{#1}.} 
\DeclareMathOperator*{\argmax}{argmax}
\DeclareMathOperator*{\argmin}{argmin}
\newcommand{\expect}[1]{\textsf{E}\{#1\}}
\newcommand{\expectwrt}[2]{\textsf{E}_{#2}\{#1\}}

\newcommand{\ucbone}{\textsc{UCB1}}
\newcommand{\ucbonetuned}{\textsc{UCB1-Tuned}}
\newcommand{\ucbonenormal}{\textsc{UCB1-Normal}}
\newcommand{\ucbtwo}{\textsc{UCB2}}
\newcommand{\ucbv}{\textsc{UCB-V}}
\newcommand{\klucb}{\textsc{KL-UCB}}
\newcommand{\epsilongreedy}{\textsc{$\epsilon_n$\hspace{-1pt}-Greedy}}
\newcommand{\powerone}{\textsc{Power-1}}
\newcommand{\powertwo}{\textsc{Power-2}}
\newcommand{\formulaone}{\textsc{Formula-1}}
\newcommand{\formulatwo}{\textsc{Formula-2}}

\newcommand{\RewardMean}{\ensuremath{\overline{r}_k}}
\newcommand{\RewardStddev}{\ensuremath{\overline{\sigma}_k}}
\newcommand{\PlayedCount}{\ensuremath{t_k}}
\newcommand{\TimeStep}{\ensuremath{t}}

\newcommand{\Formula}{\ensuremath{f}}

\newcommand{\Realspace}{\ensuremath{\mathbbm{R}}}
\newcommand{\dotproduct}[2]{\ensuremath{\left\langle#1, #2 \right\rangle}}

\newcommand{\formulas}{\ensuremath{\mathbb F}}
\begin{document}

\mainmatter  % start of an individual contribution

\title{Meta-Learning of Exploration/Exploitation Strategies: 
  The Multi-Armed Bandit Case}

\titlerunning{Meta-Learning of Exploration/Exploitation Strategies}

\author{Francis Maes\and Louis Wehenkel\and Damien Ernst}
\authorrunning{Francis Maes\and Louis Wehenkel\and Damien Ernst}

\institute{ University of Li\`ege \\ Dept. of Electrical Engineering and Computer Science \\
Institut Montefiore, B28, B-4000, Li\`ege - Belgium
\mailsc }


\toctitle{-}
\tocauthor{-}
\maketitle

\begin{abstract}
The exploration/exploitation (E/E) dilemma arises naturally in many subfields of Science. Multi-armed bandit problems formalize this dilemma in its canonical form. Most current research in this field focuses on generic solutions that can be applied to a wide range of problems. However, in practice, it is often the case that a form of prior information is available about the specific  class of target problems. Prior knowledge is rarely used in current solutions due to the lack of a systematic approach to incorporate it into the E/E strategy. 

To address a specific class of E/E problems, we propose to proceed in three steps: (i) model prior knowledge in the form of a probability distribution over the target class of E/E problems; (ii) choose a large hypothesis space of candidate E/E strategies; and (iii), solve an optimization problem to find a candidate E/E strategy of maximal average performance over a sample of problems drawn from the prior distribution. 

We illustrate this meta-learning approach with two different hypothesis spaces: one where E/E strategies are numerically parameterized and another where E/E strategies are represented as small symbolic formulas. We propose appropriate optimization algorithms for both cases. Our experiments, with two-armed ``Bernoulli'' bandit problems and various playing budgets, show that the meta-learnt E/E strategies  outperform  generic strategies of the literature (\ucbone , \ucbonetuned , \ucbv , \klucb\ and \epsilongreedy ); they also evaluate the robustness of the learnt E/E strategies, by tests carried out on arms whose rewards follow a truncated Gaussian distribution.
\keywords{exploration-exploitation dilemma, prior knowledge, multi-armed bandit problems, reinforcement learning}
\end{abstract}

\section{Introduction}

Exploration versus exploitation (E/E) dilemmas arise in many sub-fields of Science, and in related fields such as artificial intelligence, finance, medicine and engineering. In its most simple version, the multi-armed bandit problem formalizes this dilemma as follows \cite{Robbins52sequential}: a gambler has $T$ coins, and at each step he may choose among one of $K$ slots 
(or arms) to allocate one of these coins, and then earns some money (his reward) depending on the response of the machine he selected. Each arm response is characterized by an unknown probability distribution that is constant over time. The goal of the gambler is to collect the largest cumulated reward once he has exhausted his coins (i.e. after $T$ plays). A rational (and risk-neutral) gambler knowing the reward distributions of the $K$ arms would play at every stage an arm with maximal expected reward, so as to maximize his expected cumulative reward (irrespectively of the number $K$ of arms, his number $T$ of coins, and the variances of the reward distributions). When reward distributions are unknown, it is less trivial to decide how to play optimally since two contradictory goals compete: \textit{exploration} consists in trying an arm to acquire knowledge on its expected reward, while \textit{exploitation} consists in using the current knowledge to decide which arm to play. How to balance the effort towards these two goals is the essence of the E/E dilemma, which is specially difficult when imposing a finite number of playing opportunities $T$.

Most theoretical works about multi-armed bandit problem have focused on the design of generic E/E strategies which are provably optimal in asymptotic conditions (large $T$), while assuming only very unrestrictive conditions on the reward distributions (e.g.,  bounded support). Among these, some strategies work by computing at every play a quantity called ``upper confidence index'' for each arm that depends on the rewards collected so far by this arm, and by selecting for the next play (or round of plays) the arm with the highest index. Such E/E strategies are called \textit{index-based policies} and have been initially introduced by \cite{Lai85asymptot} where the indices were difficult to compute. More easy to compute indices where proposed  later on \cite{Agrawal95bandits,Auer02bandits,Audibert07bandits}.

Index-based policies typically involve hyper-parameters whose values impact their relative performances. Usually, when reporting simulation results, authors manually tuned these values on problems that share similarities with their test problems (e.g., the same type of distributions for generating the rewards) by running trial-and-error simulations \cite{Auer02bandits,Audibert08variance}. By doing so, they actually used prior information on the problems to select the hyper-parameters.

Starting from these observations, we elaborated an approach for learning in a reproducible way good policies for playing multi-armed bandit problems over finite horizons. This approach explicitly models and then exploits the prior information on the target set of multi-armed bandit problems. We assume that this prior knowledge is represented as a distribution over multi-armed bandit problems, from which we can draw any number of training problems. Given this distribution, meta-learning consists in searching in a chosen set of candidate E/E strategies one that yields optimal expected performances. This approach allows to automatically tune hyper-parameters of existing index-based policies. But, more importantly, it opens the door for  searching within much broader classes of E/E strategies one that is optimal for a given set of problems compliant with  the prior information. We propose two such hypothesis spaces composed of index-based policies: in the first one, the index function is a linear function of features and whose meta-learnt parameters are real numbers, while in the second one it is a  function generated by a grammar of symbolic formulas. 

We empirically show, in the case of Bernoulli arms, that when the number $K$ of arms and the playing horizon $T$ are fully specified a priori, learning enables to obtain policies that significantly outperform a wide range of previously proposed generic policies (\ucbone , \ucbonetuned ,  \ucbtwo , \ucbv , \klucb\ and \epsilongreedy ), even after careful tuning. We also evaluate the robustness of the  learned policies with respect to erroneous prior assumptions, by testing the E/E strategies learnt for Bernoulli arms on bandits with rewards following a truncated Gaussian distribution.

The ideas presented in this paper take their roots in two previously published papers. The idea of learning multi-armed bandit policies using global optimization and numerically parameterized index-based policies was first proposed in \cite{Maes2012Icaart}. Searching good multi-armed bandit policies in a formula space was first proposed in \cite{Maes11EWRLb}. Compared to this previous work, we adopt here a unifying perspective, which is the learning of E/E strategies from prior knowledge. We also introduce an improved optimization procedure for formula search, based on equivalence classes identification and on a pure exploration multi-armed problem formalization.

This paper is structured as follows. We first formally define the multi-armed bandit problem and introduce index-based policies in Section 2. Section 3 formally states of E/E strategy learning problem. Section 4 and Section 5 present the numerical and symbolic instantiation of our learning approach, respectively. Section 6 reports on experimental results. Finally, we conclude and present future research directions in Section 7.

\section{Multi-armed bandit problem and policies}

We now formally describe the (discrete) multi-armed bandit problem and the class of index-based policies.

\subsection{The multi-armed bandit problem}

We denote  by $i \in \{1,2,\ldots,K\}$ the ($K \ge 2$) arms of the bandit problem,   by $\nu_i$ the reward distribution of arm $i$, and by $\mu_i$ its expected value;  $b_t$ is the arm played at round $t$,  and $r_t \sim\nu_{b_t}$ is the obtained reward. $H_t = [b_1, r_1, b_2, r_2, \ldots, b_t, r_t]$ is a vector that gathers the history over the first $t$ plays, and we denote by $\mathcal{H}$ the set of all possible histories of any length $t$. An E/E strategy (or policy) $\pi : \mathcal{H} \rightarrow \{1,2,\ldots,K\}$ is an algorithm that processes at play $t$ the vector $H_{t-1}$ to select the arm $b_t \in \{1,2,\ldots,K\}$: $b_t = \pi(H_{t-1})$.

The regret of the policy $\pi$ after $T$ plays  is defined by:
$
R^\pi_T= T \mu^* - \sum_{t=1}^T r_t,
$
where $\mu^* = \max_{k} {\mu_k}$ refers to the expected reward of the optimal arm. The expected value of the regret represents the expected loss due to the fact that the policy does not always play the best machine. It can be written as:
 \begin{eqnarray}
\expect{R_T^\pi} = \sum_{k=1}^K \expect{T_k(T)} (\mu^* - \mu_k)\ ,
\end{eqnarray}
where $T_k(T)$ denotes the number of times the policy has drawn arm $k$ on the first $T$ rounds.

The multi-armed bandit problem aims at finding a policy $\pi^{*}$ that for a given $K$ minimizes the expected regret (or, in other words, maximizes the expected reward), ideally for  any $T$ and any $\{\nu_{i}\}_{i=1}^{K}$.

\subsection{Index-based bandit policies}
\label{ssec:indexbasedpolicies}

\begin{algorithm}[tb]
  \alglanguage{pseudocode}
   \begin{algorithmic}[1]
    \item Given scoring function $index : \mathcal{H} \times  \{1,2,\ldots,K\} \rightarrow \Realspace$,
%     \medskip
        
     \For{$t = 1$ to $K$} 
       \State Play bandit $b_t = t$  \Comment{Initialization: play each bandit once}
       \State Observe reward $r_t$
     \EndFor
     
     \For{$t=K$ to $T$}
        \State Play bandit $b_t = \argmax_{k \in  \{1,2,\ldots,K\}} index (H^{k}_{t-1}, t) $ 
       \State Observe reward $r_t$
      \EndFor
   \end{algorithmic}
   \caption{Generic index-based discrete bandit policy}
   \label{alg:GenericIndexBasedPolicy}
\end{algorithm}

Index-based bandit policies are based on a ranking  $index$ that computes for each arm $k$ a numerical value based on the sub-history of responses $H^{k}_{t-1}$ of that arm gathered at time $t$. These policies are sketched in Algorithm \ref{alg:GenericIndexBasedPolicy} and work as follows. During the first $K$ plays, they play sequentially the machines $1,2, \ldots, K$ to perform initialization. In all subsequent plays, these policies compute for every machine $k$ the score $index(H^{k}_{t-1}, t) \in \Realspace$ that depends on the observed sub-history $H^{k}_{t-1}$ of arm $k$ and possibly on $t$. At each step $t$, the arm with the largest score is selected (ties are broken at random). 

Here are some examples of popular index functions:
\begin{align}
  index^{\ucbone}(H^{k}_{t-1}, t) &= \RewardMean + \sqrt{\frac{C \ln \TimeStep}{\PlayedCount}}  \\
  index^{\ucbonetuned}(H^{k}_{t-1}, t) &= \RewardMean + \sqrt{\frac{\ln \TimeStep}{\PlayedCount} \min \big(1/4, \RewardStddev + \sqrt{\frac{2 \ln \TimeStep}{\PlayedCount}}\big)} \\
  index^{\ucbonenormal}(H^{k}_{t-1}, t) &= \RewardMean + \sqrt{16 \frac{\PlayedCount \RewardStddev^2}{\PlayedCount - 1} \frac{\ln(\TimeStep - 1)}{\PlayedCount}} \\
  index^{\ucbv}(H^{k}_{t-1}, t) &= \RewardMean + \sqrt{\frac{2 \RewardStddev^2 \zeta \ln \TimeStep }{\PlayedCount}} + c \frac{3\zeta \ln \TimeStep }{\PlayedCount}
\end{align}
where $\RewardMean$ and $\RewardStddev$ are the mean and standard deviation of the rewards  so far obtained from arm $k$ and $\PlayedCount$ is the number of times it has been played.

Policies \ucbone , \ucbonetuned\ and \ucbonenormal \footnote{Note that this index-based policy does not strictly fit inside Algorithm \ref{alg:GenericIndexBasedPolicy} as it uses an additional condition to play bandits that were not played since a long time.} have been proposed by \cite{Auer02bandits}. \ucbone\ has one parameter $C > 0$ whose typical value is 2. Policy $\ucbv$ has been proposed by \cite{Audibert07bandits} and has two parameters $\zeta > 0$ and $c > 0$. We refer the reader to \cite{Auer02bandits,Audibert07bandits} for detailed explanations of these parameters. 
Note that these index function are the sum of an exploitation term to give preference on arms with high reward mean ($\RewardMean$)  and an exploration term that aims at playing arms to gather more information on their underlying reward distribution (which is typically an upper confidence term).

\section{Learning exploration/exploitation strategies}
\label{sec:contribution}

Instead of relying on a fixed E/E strategy to solve a given class of problems, we propose a systematic approach to exploit prior knowledge by learning E/E strategies in a problem-driven way. We now state our learning approach in abstract terms. 

Prior knowledge is represented as a distribution $\mathcal{D}_P$ over bandit problems $P = (\nu_1, \dots, \nu_K)$. From this distribution, we can sample as many training problems as desired. In order to learn E/E strategies exploiting this knowledge, we rely on a parametric family of candidate strategies $\Pi_\Theta \subset  \{1,2,\ldots,K\}^{\mathcal{H}}$ whose members are policies $\pi_\theta$ that are fully defined given parameters $\theta \in \Theta$. Given $\Pi_\Theta$, the learning problem aims at solving:
\begin{eqnarray}
\label{eq:theoreticalProblem}
  \theta^* = \argmin_{\theta \in \Theta} \expectwrt{\expect{R_{P,T}^\pi}}{P \sim \mathcal{D}_P}\ , 
\end{eqnarray}
where $\expect{R_{P,T}^\pi}$ is the expected cumulative regret of $\pi$ on problem $P$ and where $T$ is the (a-priori given) time playing horizon. Solving this minimization problem is non trivial since it involves an expectation over an infinite number of problems. Furthermore, given a problem $P$, computing $\expect{R^\pi_{P,T}}$ relies on the expected values of $T_k(T)$, which we cannot compute  exactly in the general case. Therefore, we propose to approximate the expected cumulative regret by the empirical mean regret over a finite set of training problems $P^{(1)}, \dots, P^{(N)}$ from $\mathcal{D}_P$:
\begin{eqnarray}
\label{eq:optimizationProblem}
  \theta^* = \argmin_{\theta \in \Theta} \Delta(\pi_\theta) \mbox{ where }
 \Delta(\pi) = \frac{1}{N} \sum_{i=1}^N R_{P^{(i)},T}^\pi\  ,
\end{eqnarray}
and where $R_{P^{(i)},T}^{\pi_\theta}$ values are estimated performing a single trajectory of $\pi_\theta$ on problem $P$. Note that the number of training problems $N$ will typically be large in order to make the variance $\Delta(\cdot)$ reasonably small.

In order to instantiate this approach, two components have to be provided: the hypothesis space $\Pi_\Theta$ and the optimization algorithm to solve Eq.~\ref{eq:optimizationProblem}. The next two sections describe different instantiations of these components.

\section{Numeric parameterization}

We now instantiate our meta-learning approach by considering E/E strategies that have numerical parameters.

\subsection{Policy search space}
\label{ssec:numericalspace}

To define the parametric family of candidate policies $\Pi_\Theta$, we use index functions expressed as linear combinations of history features. These index functions rely on an \textit{history feature function}  $\phi : \mathcal{H} \times  \{1,2,\ldots,K\} \rightarrow \Realspace^d$, that describes the history \textit{w.r.t.} a given arm as a vector of scalar features. Given the function $\phi(\cdot, \cdot)$, index functions are defined by
$$ index_\theta(H_t, k) =  \dotproduct{\theta}{\phi(H_t, k)},$$
where $\theta \in \Realspace^d$ are parameters and $\dotproduct{\cdot}{\cdot}$ is the classical dot product operator. The set of candidate policies $\Pi_\Theta$ is composed of all index-based policies obtained with such index functions given parameters $\theta \in \Realspace^d$.

History features may describe any aspect of the history, including empirical reward moments, current time step, arm play counts or combinations of these variables. The set of such features should not be too large to avoid parameter estimation difficulties, but it should be large enough to provide the support for a rich set of E/E strategies. We here propose one possibility for defining the history feature function, that can be applied to any multi-armed problem and that is shown to perform well in Section \ref{sec:experiments}.

To compute $\phi(H_t, k)$, we first compute the following four variables: $v_1 = \sqrt{\ln \TimeStep}, v_2 = 1 / \sqrt{\PlayedCount}, v_3 =  \RewardMean$ and $v_4 = \RewardStddev$, i.e. the square root of the logarithm of the current time step, the inverse square root of the number of times arm $k$ has been played, the empirical mean and standard deviation of the rewards obtained so far by arm $k$.

%$v_1 = \sqrt{\ln t}$ is the square root of the logarithm of the current time step, $v_2 = \sqrt{\frac{1}{T_k}}$ is the inverse of the square root of the play count of machine $k$, $v_3 =  \bar{r}_k$ is the average reward obtained from arm $k$ so far and $v_4 = \bar{\rho_k}$ is the empirical standard deviation of these rewards. 

Then, these variables are multiplied in different ways to produce features. The number of these combinations is controlled by a parameter $P$ whose default value is $1$. Given $P$, there is one feature $f_{i,j,k,l}$ per possible combinations of values of $i,j,k,l \in \{0,\ldots, P\}$, which is defined as follows: $ f_{i,j,k,l} = v_1^i v_2^j v_3^k v_4^l$.

In other terms, there is one feature per possible polynomial up to degree $P$ using variables $v_1, \dots, v_4$. In the following, we denote  $\powerone$  (resp., $\powertwo$) the policy learned using function $\phi(H_t, k)$ with parameter $P=1$ (resp., $P=2$). The index function that underlies these policies can 
be written as following:
\begin{eqnarray}
  index^{power-P}(H_t, k) = \sum_{i=0}^{P}\sum_{j=0}^{P}\sum_{k=0}^{P}\sum_{l=0}^{P} \theta_{i,j,k,l} v_1^i v_2^j v_3^k v_4^l
\end{eqnarray}
where $\theta_{i,j,k,l}$ are the learned parameters. The $\powerone$ policy has $16$ such parameters and the $\powertwo$ has $81$ parameters.

\subsection{Optimisation algorithm}

We now discuss the optimization of Equation \ref{eq:optimizationProblem} in the case of our numerical parameterization. Note that the objective function we want to optimize, in addition to being stochastic, has a complex relation with the parameters $\theta$. A slight change in the parameter vector $\theta$ may lead to significantly different bandit episodes and expected regret values. Local optimization approaches may thus not be appropriate here. Instead, we suggest the use of derivative-free global optimization algorithms.

In this work, we use a powerful, yet simple, class of global optimization algorithms known as \textit{cross-entropy} and also known as \textit{Estimation of Distribution Algorithms} (EDA) \cite{lozano02eda}. EDAs rely on a probabilistic model to describe promising regions of the search space and to sample good candidate solutions. This is performed by repeating iterations that first \textit{sample} a population of $n_{p}$ candidates using the \textit{current} probabilistic model and then \textit{fit} a \textit{new} probabilistic model given the $b < n_{p}$ best candidates. 

\begin{algorithm}[tb]
  \alglanguage{pseudocode}
   \begin{algorithmic}[1]
    \item[\textit{Given}] the number of iterations $i_{max}$,
    \item[\textit{Given}] the population size $n_{p}$,
    \item[\textit{Given}] the number of best elements $b$,
    \item[\textit{Given}] a sample of training bandit problems $P^{(1)}, \dots, P^{(N)}$,
    \item[\textit{Given}] an history-features function $\phi(\cdot,\cdot) \in \Realspace^d$,
    \medskip

    \State Set $\mu_p = 0, \sigma^2_p = 1\ , \forall p \in [1,d]$ \Comment{Initialize with normal Gaussians}
     
     \For{$i \in [1, i_{max}]$}
     \smallskip
     \For{$j \in [1,n_{p}]$}      \Comment{Sample and evaluate new population}
        \For{$p \in [1,d]$} 
           \State $\theta_p \leftarrow $ sample from $\mathcal{N}(\mu_p, \sigma_p^2)$
        \EndFor
        \State Estimate $\Delta(\pi_\theta)$ and store result $(\theta, \Delta(\pi_\theta))$
%        \If{$\Delta(\pi_\theta)$ is lower than any previously observed score}
%          \State $\theta^* = \theta$
%        \EndIf
     \EndFor

     \smallskip
        
     \State Select $\{\theta^{(1)}, \dots, \theta^{(b)} \}$ the $b$ best candidate $\theta$ vectors \textit{w.r.t.} their $\Delta(\cdot)$ score 

     \smallskip
       \State $\mu_p \leftarrow  \frac{1}{b} \sum_{j=1}^{b} \theta^{(j)}_p \ ,  \forall p \in[1,d]$  \Comment{Learn new Gaussians}
       \State $\sigma^2_p \leftarrow \frac{1}{b} \sum_{j=1}^{b} (\theta^{(j)}_p - \mu_p)^2 \ , \forall p \in[1,d]$ 
    \EndFor
     \State \Return{ The policy $\pi_{\theta}$ that led to the lowest observed value of $\Delta(\pi_\theta)$ } 
  \end{algorithmic}
\caption{EDA-based learning of a discrete bandit policy}
\label{alg:EDA}
\end{algorithm}

Any kind of probabilistic model may be used inside an EDA. The simplest form of EDAs uses one marginal distribution per variable to optimize and is known as the \textit{univariate marginal distribution algorithm} \cite{Pelikan98marginaleda}.  We have adopted this approach by using one Gaussian distribution $\mathcal{N}(\mu_p, \sigma_p^2)$ for each parameter $\theta_p$. Although this approach is simple, it proved to be quite effective experimentally to solve Equation \ref{eq:optimizationProblem}.
% With this model, probability density functions of $\theta \in \mathbbm{R}^p$ vectors have the following form:
%$$
%   f(\theta) = \prod_{f=1}^{p} \frac{1}{\sqrt{2\pi\sigma_p^2}} {e^{-\frac{(x-\mu_p)^2}{2\sigma_p^2} }}
%$$
The full details of our EDA-based policy learning procedure are given by Algorithm \ref{alg:EDA}. 
The initial distributions are standard Gaussian distributions $\mathcal{N}(0,1)$. The policy that is returned corresponds to the $\theta$ parameters that led to the lowest observed value of $\Delta(\pi_\theta)$.

\section{Symbolic parametrization}

The index functions from the literature depend on the current time step $t$ and on  three statistics extracted from the sub-history $H^{k}_{t-1}$ : $\RewardMean$, $\RewardStddev$ and $\PlayedCount$. We now propose a second parameterization of our learning approach, in which we consider all index functions that can be constructed using small formulas built upon these four variables.

\subsection{Policy search space}

We consider  index functions that are given in the form of small, closed-form formulas. Closed-form formulas have several advantages: they can be easily computed, they can formally be analyzed and they are easily interpretable.

Let us first explicit the set of formulas $\formulas$  that we consider in this paper. A formula $F \in \formulas$ is: 
\begin{itemize}
\item either a binary expression $F = B(F',F'')$, where $B$ belongs to a set of binary operators $\mathbb B$ and $F'$ and $F''$ are also formulas from $\formulas$,
\item or a unary expression $F = U(F')$ where $U$ belongs to a set of unary operators $\mathbb U$ and $F' \in \formulas$, 
\item or an atomic variable $F = V$, where $V$ belongs to a set of variables  $\mathbb V$,
\item  or a constant $F = C$, where $C$ belongs to a set of constants $\mathbb C$. 
\end{itemize}

In the following, we consider a set of operators and constants that provides a good compromise between high expressiveness  and low cardinality of $\formulas$.
The set of binary operators considered in this paper $\mathbb B$ includes the four elementary mathematic operations and the $\min$ and $\max$ operators: $\mathbb B = \left\{ +,  - ,  \times , \div , \min, \max  \right\}$.
The set of unary operators $\mathbb U$ contains the square root, the logarithm, the absolute value, the opposite and the inverse:
$\mathbb U = \left\{  \sqrt{. }, \ln(.), | . |, -., \frac{1}{.}   \right\} $.
The set of variables $\mathbb V$ is:
$\mathbb V = \left\{ \RewardMean, \RewardStddev, \PlayedCount, \TimeStep \right\}$.
The set of  constants $\mathbb C$ has been chosen to maximize the number of different numbers representable by small formulas. It is defined as $\mathbb C = \{1,2,3,5,7 \}$.

\begin{figure}[tb]
\begin{minipage}[b]{0.4\linewidth}
\begin{align*}
   F & ::= B(F,F) \mbox{ $|$ } U(F) \mbox{ $|$ } V \mbox{ $|$ } C \\
   B & ::= +\mbox{ $|$ }  -\mbox{ $|$ }  \times \mbox{ $|$ } \div \mbox{ $|$ }  min \mbox{ $|$ }  max  \\
   U & ::=  sqrt \mbox{ $|$ }  ln \mbox{ $|$ } abs  \mbox{ $|$ } opposite \mbox{ $|$ } inverse \\
   V & ::= \RewardMean \mbox{ $|$ } \RewardStddev \mbox{ $|$ } \PlayedCount \mbox{ $|$ }  \TimeStep \\
   C & ::= 1,2,3,5,7
\end{align*}
%\vspace{3pt}
\end{minipage}
\begin{minipage}[b]{0.65\linewidth}

\includegraphics[width=0.8\textwidth]{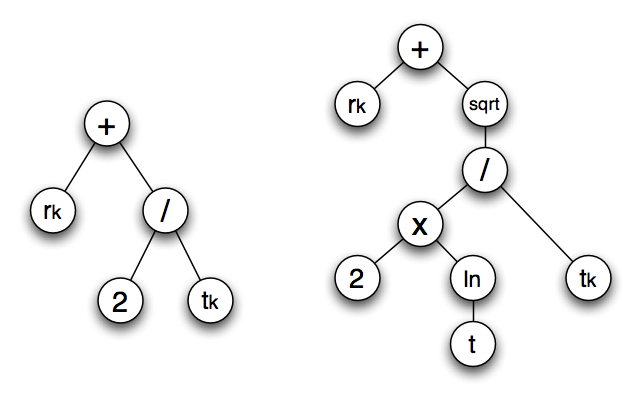}
\vspace{-10pt}
\end{minipage}
\caption{The grammar used for generating candidate index functions and two example formula parse trees corresponding to $\RewardMean + 2 / \PlayedCount$ and $\RewardMean + \sqrt{2 ln (\TimeStep) / \PlayedCount}$. }
\label{fig:grammar}
\end{figure}

Figure \ref{fig:grammar} summarizes our grammar of formulas and gives two examples of index functions.  The length of a formula $length(\Formula)$ is the number of symbols occurring in the formula. For example, the length of $\RewardMean + 2 / \PlayedCount$ is 5 and the length of $\RewardMean + \sqrt{2 \times ln (\TimeStep) / \PlayedCount}$ is 9. Let $L$ be a given maximal length. $\Theta$ is the subset of formulas whose length is no more than $L$:
$
  \Theta = \{\Formula |Êlength(\Formula) \leq L\}
$ 
and $\Pi_\Theta$ is the set of index-based policies whose index functions are defined by formulas $\Formula \in \Theta$.

\subsection{Optimisation algorithm}
\label{ssec:symbOpti}

We now discuss the optimization of Equation \ref{eq:optimizationProblem} in the case of our symbolic parameterization. First, notice that several different formulas can lead to the same policy. For example, any increasing function of $\RewardMean$ defines the greedy policy, which always selects the arm that is believed to be the best. Examples of such functions in our formula search space include $\RewardMean$, $\RewardMean \times 2$, $\RewardMean \times \RewardMean$ or $\sqrt{\RewardMean}$. 

Since it is useless to evaluate equivalent policies multiple times, we propose the following two-step approach. First, the set $\Theta$ is partitioned into equivalence classes, two formulas being equivalent if and only if they lead to the same policy. Then, Equation \ref{eq:optimizationProblem} is solved over the set of equivalence classes (which is typically one or two orders of magnitude smaller than the initial set $\Theta$).

\smallparagraph{Partitioning $\Theta$} This task is far from trivial: given a formula, equivalent formulas can be obtained through commutativity, associativity, operator-specific rules and through any increasing transformation. Performing this step exactly involves advanced static analysis of the formulas, which we believe to be a very difficult solution to implement. Instead, we propose a simple approximated solution, which consists in  discriminating formulas by comparing how they rank (in terms of values returned by the formula) a set of $d$ random samples of the variables  $\RewardMean, \RewardStddev, \PlayedCount, \TimeStep$.  More formally, the procedure is the following: 

\newcommand{\bandits}{\ensuremath{M}}
\newcommand{\approxshortformulas}{\ensuremath{\tilde{\Theta}}}

\begin{enumerate}
\item we first build $\Theta$, the space of all formulas $f$ such that $length(f) \leq L$;

\item for $i=1 \ldots d$, we uniformly draw (within their respective domains)  some random realizations of the variables  $\RewardMean, \RewardStddev, \PlayedCount, \TimeStep$ that we concatenate into a vector  $\Theta_i$;

\item we cluster all formulas from $\Theta$ according to the following rule: two formulas $F$ and $F'$ belong to the same cluster if and only if they rank all the $\Theta_i$ points in the same order, i.e.: $\forall i, j \in \{1, \ldots, d \}, i\neq j, F(\Theta_i) \geq  \ F(\Theta_j) \iff F'(\Theta_i) \geq  \ F'(\Theta_j)$.
Formulas leading to invalid index functions  (caused for instance by division by zero or logarithm of negative values) are discarded; 

\item among each cluster, we select one formula of minimal length;

\item we gather all the selected  minimal length formulas into an approximated reduced set of formulas $\approxshortformulas$.
\end{enumerate}

 In the following, we denote by $\bandits$ the cardinality of the approximate set of formulas $\approxshortformulas = \left\{ f_1, \ldots, f_\bandits  \right\}$.

\medskip

% % % % % % % % % % % % % % % % % % % 
\smallparagraph{Optimization algorithm} A naive approach for finding the best formula  ${\Formula^*} \in \approxshortformulas$ would be to evaluate $\Delta(\Formula)$ for each formula $\Formula \in \approxshortformulas$ and simply return the best one. While extremely simple to implement, such an approach could reveal itself to be time-inefficient in case of  spaces $\approxshortformulas$ of large cardinality.

Preliminary experiments have shown us that $\approxshortformulas$ contains a majority of formulas leading to relatively bad performing index-based policies. It turns out that relatively few samples of $R_{P^{(i)},T}^\pi$ are sufficient to reject with high confidence these badly performing formulas. In order to exploit this idea, a natural idea is to formalize the search for the best formula as another multi-armed bandit problem. To each formula  $F_k \in \approxshortformulas$, we associate an arm. Pulling the arm $k$ consists in selecting a training problem $P^{(i)}$ and in running one episode with the index-based policy whose index formula is $\Formula_k$. This leads to a reward associated to arm $k$ whose value is the quantity $-R_{P^{(i)},T}^\pi$ observed during the episode. The purpose of multi-armed bandit algorithms is here to process the sequence of observed rewards to select in a smart way the next formula to be tried so that when the budget of pulls has been exhausted, one (or several) high-quality formula(s) can be identified. 

In the formalization of Equation~\ref{eq:optimizationProblem} as a multi-armed bandit problem, only the quality of the finally suggested arm matters. How to select arms so as to identify the best one in a finite amount of time is known as the \textit{pure exploration} multi-armed bandit problem \cite{Bubeck09pure}.  
It has been shown that index-based policies based on upper confidence bounds were good policies for solving pure exploration bandit problems. Our optimization procedure works as follows: we use a bandit algorithm such as \ucbonetuned\ during a given number of steps and then return the policy that corresponds to the formula $\Formula_k$ with highest expected reward $\RewardMean$. The problem instances are selected depending on the number of times the arm has been played so far: at each step, we select the training problem $P^{(i)}$ with $i = 1 + (\PlayedCount\mbox{ mod }N)$. 

In our experiments, we estimate that our multi-armed bandit approach is one hundred to one thousand times faster than the naive Monte Carlo optimization procedure, which clearly demonstrates the benefits of this approach. Note that this idea could also be relevant to our numerical case. The main difference is that the corresponding multi-armed bandit problem relies on a continuous-arm space. Although some algorithms have already been proposed to solve such multi-armed bandit problems \cite{Bubeck2011XArmed}, how to scale these techniques to problems with hundreds or thousands parameters is still an open research question. Progresses in this field could directly benefit our numerical learning approach.

\section{Numerical experiments}
\label{sec:experiments}

We now illustrate the two instances of our learning approach by comparing learned policies against a number of generic previously proposed policies in a setting where prior knowledge is available about the target problems. We show that in both cases, learning enables to obtain exploration/exploitation strategies significantly outperforming all tested generic policies. 

\subsection{Experimental protocol}

We compare learned policies against generic policies. We distinguish between \textit{untuned generic policies} and  \textit{tuned generic policies}. The former are either policies that are parameter-free or policies used with default parameters suggested in the literature, while the latter are generic policies whose hyper-parameters were tuned using Algorithm \ref{alg:EDA}.

\smallparagraph{Training and testing} To illustrate our approach, we consider the scenario where the number of arms $K$, the playing horizon $T$ and the kind of distributions $\nu_k$ are known a priori 
and where the parameters of these distributions are missing information. Since we are learning policies, care should be taken with generalization issues. As usual in supervised machine learning, we use a training set which is distinct from the testing set. The training set is composed of $N=100$ bandit problems sampled from a given distribution over bandit problems $\mathcal{D}_P$ whereas the testing set contains another $10000$ problems drawn from this distribution. To study the robustness of our policies w.r.t. wrong prior information, we also report their performance on a set of $10000$ problems drawn from another distribution $\mathcal{D}^\prime_P$ with different kinds of distributions $\nu_k$. When computing $\Delta(\pi_\theta)$, we estimate the regret for each of these problems by averaging results overs $100$ runs. One calculation of $\Delta(\pi_\theta)$ thus involves simulating $10^4$ (resp. $10^6$) bandit episodes during training (resp. testing). 

\smallparagraph{Problem distributions} The distribution $\mathcal{D}_P$ is composed of two-armed bandit problems with Bernoulli distributions whose expectations are uniformly drawn from  $[0,1]$. Hence, in order to sample a bandit problem from $\mathcal{D}_P$, we draw the expectations $p_1$ and $p_2$ uniformly from $[0,1]$ and return the bandit problem with two Bernoulli arms that have expectations $p_1$ and $p_2$, respectively. In the second distribution  $\mathcal{D}^\prime_P$, the reward distributions $\nu_k$ are changed by Gaussian distributions truncated to the interval $[0,1]$. In order to sample one problem from $\mathcal{D}^\prime_P$, we select a mean and a standard deviation for each arm uniformly in range $[0,1]$. Rewards are then sampled using a rejection sampling approach: samples are drawn from the corresponding Gaussian distribution until obtaining a value that belongs to the interval $[0,1]$.

\smallparagraph{Generic policies} We consider the following generic policies: the \epsilongreedy\ policy  as described in \cite{Auer02bandits}, the policies introduced by \cite{Auer02bandits}: \ucbone , \ucbonetuned , \ucbonenormal\ and \ucbtwo, the policy \klucb\ introduced in \cite{garivier2011klucb}  and the policy \ucbv\ proposed by \cite{Audibert07bandits}. Except \epsilongreedy , all these policies belong to the family of index-based policies discussed previously. \ucbonetuned\ and \ucbonenormal\ are parameter-free policies designed for bandit problems with Bernoulli distributions and for problems with Gaussian distributions respectively. All the other policies have hyper-parameters that can be tuned to improve the quality of the policy. \epsilongreedy\ has two parameters $c > 0$ and $0 < d < 1$, \ucbtwo\ has one parameter $0 < \alpha < 1$, $\klucb$ has one parameter $c \geq 0$ and $\ucbv$ has two parameters $\zeta > 0$ and $c > 0$. We refer the reader to \cite{Auer02bandits,Audibert07bandits,garivier2011klucb} for detailed explanations of these parameters. 

\smallparagraph{Learning numerical policies} We learn policies using the two parameterizations \powerone\ and \powertwo\ described in Section \ref{ssec:numericalspace}. Note that tuning generic policies is a particular case of learning with numerical parameters and that both learned policies and tuned generic policies make use of the same prior knowledge. To make our comparison between these two kinds of policies fair, we always use the same training procedure, which is Algorithm \ref{alg:EDA} with $i_{max} = 100$ iterations, $n_{p} = max(8 d,  40)$ candidate policies per iteration and $b = n_{p}/ 4$ best elements, where $d$ is the number of parameters to optimize. Having a linear dependency between $n_{p}$ and $d$ is a classical choice when using EDAs \cite{Rubenstein04CE}. Note that, in most cases the optimization is solved in a few or a few tens iterations. Our simulations have shown that $i_{max} = 100$ is a careful choice for ensuring that the optimization has enough time to properly converge. For the baseline policies where some default values are advocated, we use these values as initial expectation of the EDA Gaussians. Otherwise, the initial Gaussians are centered on zero. Nothing is done to enforce the EDA to respect the constraints on the parameters (e.g., $c > 0$ and $0 < d < 1$ for \epsilongreedy ). In practice, the EDA automatically identifies interesting regions of the search space that respect these constraints.

\smallparagraph{Learning symbolic policies} We apply our symbolic learning approach with a maximal formula length of $L = 7$, which leads to a set of $|\Theta| \approx 33,5$ millions of formulas. We have applied the approximate partitioning approach described in Section \ref{ssec:symbOpti} on these formulas using $d=1024$ samples to discriminate among strategies. This has resulted in $\approx 9,5$ million invalid formulas and $\bandits = 99 020$ distinct candidate E/E strategies (i.e. distinct formula equivalence classes). To identify the best of those distinct strategies, we apply the \ucbonetuned\ algorithm for $10^7$ steps. In our experiments, we report the two best found policies, which we denote \formulaone\ and \formulatwo .

\subsection{Performance comparison}
%
%\begin{figure}[tb]
%\twocurves{results/Training1.pdf}{results/Training2.pdf}
% 
%\caption{Training and testing regret of \powerone\ as a function of the number of EDA iterations. Left: training horizon $100$. Right: training horizon $1000$.}
% \label{fig:Training}
%\end{figure} 

\begin{table}[h!]
\begin{small}
   \begin{tabular*}{\textwidth}{@{\extracolsep{\fill}} |cc|c|ccc|ccc|}
    \hline
    Policy & Training & Parameters & \multicolumn{3}{c|}{Bernoulli} & \multicolumn{3}{c|}{Gaussian} \\
     & Horizon & & \tiny{T=10} & \tiny{T=100} & \tiny{T=1000} & \tiny{T=10} & \tiny{T=100} & \tiny{T=1000} \\
    \hline
      \multicolumn{9}{c}{} \\ 
   \multicolumn{9}{c}{\textit{Untuned generic policies}} \\ 
    \hline
    \ucbone &   - & $C = 2$ & 1.07 & 5.57 & 20.1 & 1.37 & 10.6 & 66.7 \\
\hspace{-15pt}   \ucbonetuned \hspace{-15pt} & - &  & \textbf{0.75} & \textbf{2.28} & \textbf{5.43} & \textbf{1.09} & 6.62 & \textbf{37.0} \\
    \ucbonenormal &  - &  & 1.71 & 13.1 & 31.7 & 1.65 & 13.4 & 58.8 \\
    \ucbtwo & - & $\alpha = 10^{-3}$  & 0.97 & 3.13 & 7.26 & 1.28 & 7.90 & 40.1 \\
    \ucbv &  - & $c=1, \zeta=1$ & 1.45 & 8.59 & 25.5 & 1.55 & 12.3 & 63.4 \\
    \klucb & - & $c=0$ & 0.76 & 2.47 & 6.61 & 1.14 & 7.66 & 43.8 \\
    \klucb & - & $c=3$ & 0.82 & 3.29 & 9.81 & 1.21 & 8.90 & 53.0 \\
    \epsilongreedy & - & $c = 1, d = 1$ & 1.07 & 3.21 & 11.5 & 1.20 & \textbf{6.24} & 41.4 \\
     \hline
      \multicolumn{9}{c}{} \\ 
   \multicolumn{9}{c}{\textit{Tuned generic policies}} \\ 
     \hline
                    &  \scriptsize{T=10}  & $C = 0.170$ & \textit{0.74} & \textbf{2.05} & 4.85 & 1.05 & 6.05 & 32.1 \\
     \ucbone &  \scriptsize{T=100}  & $C = 0.173$ & 0.74 & \textit{\textbf{2.05}} & 4.84 & 1.05 & 6.06 & 32.3 \\
                      &  \scriptsize{T=1000} & $C = 0.187$ & 0.74 & 2.08 & \textit{\textbf{4.91}} & 1.05 & 6.17 & 33.0 \\
     \hline
                    & \scriptsize{T=10}  &$\alpha = 0.0316 $ & \textit{0.97} & 3.15 & 7.39 & 1.28 & 7.91 & 40.5 \\
     \ucbtwo & \scriptsize{T=100}  & $\alpha = 0.000749$ & 0.97 & \textit{3.12} & 7.26 & 1.33 & 8.14 & 40.4 \\
                    & \scriptsize{T=1000} & $\alpha = 0.00398$ & 0.97 & 3.13 & \textit{7.25} & 1.28 & 7.89 & 40.0 \\
     \hline 
                & \scriptsize{T=10}  & $c=1.542, \zeta=0.0631$ & \textit{0.75} & 2.36 & 5.15 & \textbf{1.01} & 5.75 & 26.8 \\
     \ucbv & \scriptsize{T=100}  & $c=1.681, \zeta =0.0347$ & 0.75 & \textit{2.28} & 7.07 & \textbf{1.01} & \textbf{5.30} & \textbf{27.4} \\
                & \scriptsize{T=1000} & $c=1.304, \zeta=0.0852$ & 0.77 & 2.43 & \textit{5.14} & 1.13 & 5.99 & 27.5 \\
     \hline 
                & \scriptsize{T=10}  & $c=-1.21$ & \textit{\textbf{0.73}} & 2.14 & 5.28 & 1.12 & 7.00 & 38.9 \\
    \klucb & \scriptsize{T=100}  &  $c=-1.82$ &  \textbf{0.73} & \textit{2.10} & 5.12 & 1.09 & 6.48 & 36.1 \\
                & \scriptsize{T=1000}  &  $c=-1.84$ &  \textbf{0.73} & 2.10 & \textit{5.12} & 1.08 & 6.34 & 35.4 \\
      \hline
                                 & \scriptsize{T=10}  & $c=0.0499, d=1.505$ & \textit{0.79} & 3.86 & 32.5 & \textbf{1.01} & 7.31 & 67.6  \\
     \epsilongreedy & \scriptsize{T=100}  & $c=1.096, d=1.349$ & 0.95 & \textit{3.19} & 14.8 & 1.12 & 6.38 & 46.6 \\
                                & \scriptsize{T=1000} & $c=0.845, d=0.738$ & 1.23 & 3.48 & \textit{9.93} & 1.32 & 6.28 & 37.7 \\
     \hline
      \multicolumn{9}{c}{} \\ 
    \multicolumn{9}{c}{\textit{Learned numerical policies}} \\ 
     \hline
                         & \scriptsize{T=10}  & $\dots$   & \textit{\textbf{0.72}} & 2.29 & 14.0 & \textbf{0.97} & 5.94 & 49.7 \\
     \powerone & \scriptsize{T=100}  & \textit{(16 parameters)} & 0.77 & \textit{1.84} & 5.64 & 1.04 & 5.13 & 27.7 \\
                        & \scriptsize{T=1000} & $\dots$ & 0.88 & 2.09 & \textit{4.04} & 1.17 & 5.95 & 28.2 \\

     \hline
                        & \scriptsize{T=10}  & $\dots$ & \textit{\textbf{0.72}} & 2.37 & 15.7 & \textbf{0.97} & 6.16 & 55.5 \\
     \powertwo & \scriptsize{T=100}  & \textit{(81 parameters)} & 0.76 & \textit{\textbf{1.82}} & 5.81 & 1.05 & \textbf{5.03} & 29.6 \\
                       & \scriptsize{T=1000} & $\dots$ &  0.83 & 2.07 & \textit{\textbf{3.95}} & 1.12 & 5.61 & \textbf{27.3} \\
     \hline
      \multicolumn{9}{c}{} \\ 
    \multicolumn{9}{c}{\textit{Learned symbolic policies}} \\ 
    \hline     
    				&  \scriptsize{T=10}  & $\sqrt{\PlayedCount} (\RewardMean - 1/2)$ &  \textit{\textbf{0.72}} & 2.37 & 14.7 	& \textbf{0.96} & 5.14 & 30.4	 \\
	\formulaone   	&  \scriptsize{T=100}  & $\RewardMean + 1 / (\PlayedCount + 1/2)$ & 0.76 & \textbf{\textit{1.85}} & 8.46 	& 1.12 & \textbf{5.07} & 29.8 \\
    				&  \scriptsize{T=1000}  & $\RewardMean + 3 / (\PlayedCount + 2)$ & 0.80 & 2.31 & \textbf{\textit{4.16}} 	& 1.23 & 6.49 & 26.4 \\
    \hline     
    				&  \scriptsize{T=10}  & $|\RewardMean - 1 / (\PlayedCount + \TimeStep)|$ &  \textit{0.72} & 2.88 & 22.8 	& 1.02 & 7.15 & 66.2 \\
	\formulatwo   	&  \scriptsize{T=100}  & $\RewardMean + min(1 / \PlayedCount, log(2))$ & 0.78 & \textit{1.92} & 6.83 	& 1.17 & 5.22 & 29.1 \\
    				&  \scriptsize{T=1000}  & $1 / \PlayedCount - 1 / (\RewardMean - 2)$ & 1.10 & 2.62 & \textit{4.29} 		& 1.38 & 6.29 & \textbf{26.1} \\
     \hline
  \end{tabular*}
  \end{small}
  \caption{Mean expected regret of untuned, tuned and learned policies on Bernoulli and Gaussian bandit problems. Best scores in each of these categories are shown in bold. Scores corresponding to policies that are tested on the same horizon $T$ than the horizon used for training/tuning are shown in italics.}
  \vspace{-40pt}
  \label{table:results1}
\end{table}

\begin{table}[bt]
  \centering
   \begin{tabular}{|c|c|c|c|c|c|c|c|}
    \hline

    Policy & T = 10 & T = 100 & T = 1000 &   Policy & T = 10 & T = 100 & T = 1000 \\
     \hline
\multicolumn{4}{c}{\textit{\small{Generic policies}}} & \multicolumn{4}{c}{\textit{\small{Learned  policies}}} \\
\hline
     \ucbone             	& 48.1 \% & 78.1 \% & 83.1 \%	& \powerone   & 54.6 \% & 82.3 \% & \textbf{91.3 \%} \\
     \ucbtwo             	& 12.7 \% & 6.8 \% & 6.8 \% 	& \powertwo   & 54.2 \% & \textbf{84.6 \%} & 90.3 \% \\
     \ucbv                  	& 38.3 \% &57.2 \% &  49.6 \% &  \formulaone	& \textbf{61.7} \% & 76.8 \% & 88.1 \% \\
     \klucb                 	& 50.5 \% & 65.0 \% & 67.0 \% & \formulatwo	& 61.0 \% & 80.0 \% & 73.1 \% \\
     \epsilongreedy 	& 37.5 \% & 14.1 \% & 10.7 \% &   & & & \\
     \hline
  \end{tabular}
  \caption{Percentage of wins against \ucbonetuned\ of generic and learned policies. Best scores are shown in bold. \vspace{-25pt}}
  
  \label{table:results2}
\end{table}

Table \ref{table:results1} reports the results we obtain for untuned generic policies, tuned generic policies and learned policies on distributions $\mathcal{D}_P$ and $\mathcal{D}^\prime_P$ with horizons $T \in \{10, 100, 1000\}$. For both tuned and learned policies, we consider three different training horizons $\{10, 100, 1000\}$ to see the effect of a mismatch between the training and the testing horizon.

\smallparagraph{Generic policies} As already pointed out in \cite{Auer02bandits}, it can be seen that \ucbonetuned\ is particularly well fitted to bandit problems with Bernoulli distributions. It also proves effective on bandit problems with Gaussian distributions, making it nearly always outperform the other untuned policies. By tuning \ucbone , we outperform the \ucbonetuned\ policy (e.g. $4.91$ instead of $5.43$ on Bernoulli problems with $T=1000$). This also sometimes happens with \ucbv . However, though we used a careful tuning procedure, \ucbtwo\ and \epsilongreedy\ do never outperform \ucbonetuned .

\smallparagraph{Learned policies} We observe that when the training horizon is the same as the testing horizon $T$, the learned policies (\powerone , \powertwo , \formulaone\ and \formulatwo ) systematically outperform all generic policies. The overall best results are obtained with \powertwo\ policies. Note that, due to their numerical nature and due to the large number of parameters, these policies are extremely hard to interpret and to understand. The results related to symbolic policies show that there exist very simple policies that perform nearly as well as these black-box policies. This clearly shows the benefits of our two hypothesis spaces: numerical policies enable to reach very high performances while symbolic policies provide interpretable strategies whose behavior can be more easily analyzed. This interpretability/performance tradeoff is common in machine learning and has been identified several decades ago in the field of supervised learning. It is worth mentioning that, among the $99 020$ formula equivalence classes, a surprisingly large number of strategies outperforming generic policies were found: when $T=100$ (resp. $T=1000$), we obtain about 50 (resp. 80) different symbolic policies outperforming the generic policies. 

\smallparagraph{Robustness \textit{w.r.t.} the horizon $T$} As expected, the learned policies give their best performance when the training and the testing horizons are equal. Policies learned with large training horizon prove to work well also on smaller horizons. However, when the testing horizon is larger than the training horizon, the quality of the policy may quickly degrade (e.g. when evaluating \powerone\  trained with $T=10$ on an horizon $T=1000$). 

\smallparagraph{Robustness \textit{w.r.t.} the kind of distribution} Although truncated Gaussian distributions are significantly different from Bernoulli distributions, the learned policies most of the time generalize well to this new setting and still outperform all the other generic policies.

\smallparagraph{A word on the learned symbolic policies} It is worth noticing that the best index-based policies (\formulaone)  found for the two largest horizons ($T=100$ and $T=1000$) work in a similar way as the UCB-type policies reported earlier in the literature.  Indeed, they also associate to an arm $k$ an index which is the sum of $\RewardMean$ and of a positive (optimistic) term that decreases with  $\PlayedCount$. However, for the shortest time horizon ($T=10$), the policy found ($\sqrt{\PlayedCount}(\RewardMean - \frac{1}{2})$) is totally different from UCB-type policies. With such a policy, only the arms whose empirical reward mean is higher than a given threshold (0.5) have positive index scores and are candidate for selection, i.e. making the scores negative has the effect to kill bad arms. If the $\RewardMean$ of an arm is above the threshold, then the index associated with this arm will increase with the number of times it is played and not decrease as it is the case for UCB policies. If all empirical means $\RewardMean$ are below the threshold, then for equal reward means, arms that have been less played are preferred. This finding is amazing since it suggests that this optimistic paradigm for multi-armed bandits upon which UCB policies are based may in fact not be adapted at all to a context where the horizon is small.

\smallparagraph{Percentage of wins against \ucbonetuned}  Table \ref{table:results2} gives for each policy, its percentage of wins against \ucbonetuned , when trained with the same horizon as the test horizon. To compute this percentage of wins, we evaluate the expected regret on each of the 10000 testing problems and count the number of problems for which the tested policy outperforms \ucbonetuned . 
We observe that by minimizing the expected regret, our learned policies  also reach high values of percentage of wins: 84.6 \% for $T=100$ and 91.3 \% for $T=1000$. Note that, in our approach, it is easy to change the objective function. So if the real applicative aim was to maximize the percentage of wins against \ucbonetuned , this criterion could have been used directly in the policy optimization stage to reach even better scores.

\subsection{Computational time}

We used a C++ based implementation to perform our experiments. In the numerical case with $10$ cores at $1.9 Ghz$, performing the whole learning  of \powerone\  took one hour for $T=100$ and ten hours for $T=1000$. In the symbolic case using a single core at $1.9Ghz$, performing the whole learning took 22 minutes for $T=100$ and a bit less than three hours for $T=1000$. Note that the fact that symbolic learning is much faster can be explained by two reasons. First, we tuned the EDA algorithm in a very careful way to be sure to find a high quality solution; what we observe is that by using only $10\%$ of this learning time, we already obtain close-to-optimal strategies. The second factor is that our symbolic learning algorithm saves a lot of CPU time by being able to rapidly reject bad strategies thanks to the multi-armed bandit formulation upon which it relies.

\section{Conclusions}
\label{sec:conclusions}

The approach proposed in this paper for exploiting prior knowledge for learning exploration/exploitation policies has been tested for two-armed bandit problems with Bernoulli reward distributions and when knowing the time horizon. The learned policies were found to significantly outperform other policies previously published in the literature  such as \ucbone , \ucbtwo , \ucbv , \klucb\ and \epsilongreedy . The robustness of the learned policies with respect to wrong information was also highlighted, by evaluating them on two-armed bandits with  truncated Gaussian reward distribution. 

 %In particular, they were still competitive with respect to these previously published policies, especially when the time horizon assumed for learning was long.

There are in our opinion several research directions that could be investigated for still improving the algorithm for learning policies proposed in this paper. For example, we found out that problems similar to the problem of overfitting met in supervised learning could occur when considering a too large set of candidate polices. This naturally calls for studying whether our learning approach could be combined with regularization techniques. Along this idea, more sophisticated optimizers could also be thought of for identifying in the set of candidate policies, the one which is predicted to behave at best.

The \ucbone , \ucbtwo , \ucbv , \klucb\ and \epsilongreedy\ policies used for comparison were shown (under certain conditions) to have interesting bounds on their expected regret in asymptotic conditions (very large $T$) while we did not aim at providing such bounds for our learned policies. It would certainly be relevant to investigate whether similar bounds could be derived for our learned policies or, alternatively,  to see how the approach could be adapted so as to target policies offering such theoretical performance guarantees in asymptotic conditions. For example, better bounds on the expected regret could perhaps be obtained by identifying in a set of candidate policies the one that gives the smallest maximal value of the expected regret over this set rather than the one that gives the best average performances.

Finally, while our paper has provided simulation results in the context of the most simple multi-armed bandit setting, our exploration/exploitation policy meta-learning scheme can also in principle be applied to any other exploration-exploitation problem. In this line of research, the extension of this investigation to (finite) Markov Decision Processes studied in \cite{Castronovo2012Ewrl}, suggests already that our approach to meta-learning E/E strategies can be successful on much more complex settings.

\bibliographystyle{splncs}
\bibliography{icaartlong}

\end{document}